%% file: ninomiya_ma_paper_2024.tex
\documentclass{amsart}
\usepackage[pdftex]{graphicx} 
\usepackage{tikz} %
\usetikzlibrary{positioning, calc, decorations.pathreplacing}
\usepackage[font=small,labelfont=bf]{caption}
\usepackage{epsfig}
\usepackage{xcolor}
\usepackage{nicematrix} 
\usepackage{amsmath,mathtools,amssymb}
\usepackage{amsthm}
\theoremstyle{plain}
\newtheorem{thm}{Theorem}
\newtheorem*{thm*}{Theorem}

\theoremstyle{definition}
\newtheorem{dfn}{Definition}
\newtheorem*{dfn*}{Definition}

\theoremstyle{remark}
\newtheorem{rem}{Remark}
\newtheorem*{rem*}{Remark}

\DeclareMathOperator*{\arginf}{arg\,inf}
\usepackage[thicklines]{cancel}

\usepackage{stackengine}
\usepackage[all]{xy} 

\begin{document}

\title[high-order deep neural SDE network]{
  {A new architecture of high-order deep neural networks
    that learn martingales
  }
}
\date{}
\author[Y.~Ma]{Yuming Ma\textsuperscript{*}}
\address{
  \textsuperscript{*}Department of Engineering and Economics,
  School of Engineering,
  Institute of Science Tokyo,
  2-12-1 Ookayama, Meguro-Ku, Tokyo 152-8552 JAPAN
}
\email{ma.y.ae@m.titech.ac.jp}
  
\author[S.~Ninomiya]{Syoiti Ninomiya\textsuperscript{\dag}}      
\address{
  \textsuperscript{\dag}Department of Mathematics,
  School of Science,
  Institute of Science Tokyo,
  2-12-1 Ookayama, Meguro-ku, Tokyo 152-8551 JAPAN
}
\email{syoiti.ninomiya@gmail.com}
\thanks{This work was supported by JSPS KAKENHI Grant Numbers 21K03365(Scientific Research (C))}

\begin{abstract}
  A new deep-learning neural network architecture based on high-order
  weak approximation algorithms for stochastic differential equations
  (SDEs) is proposed. The architecture enables the efficient learning
  of martingales by deep learning models.
  The behaviour of deep neural networks based on this architecture,
  when applied to the problem of pricing financial derivatives, is
  also examined.
  The core of this new architecture lies in the high-order weak
  approximation algorithms of the explicit Runge--Kutta type, wherein
  the approximation is realised solely through iterative compositions
  and linear combinations of vector fields of the target SDEs.
\end{abstract}
\maketitle

\section{Introduction and background}
\noindent
The objective of the present paper is to propose a new class of deep
neural network architectures for learning martingales, or, more
specifically, the vector fields that determine the stochastic
differential equations that represents these martingales.
The architecture
that we present here has
two main features.
The initial feature is that
the network learns the coefficient
functions, \mbox{i.e.} vector fields, that are contained within the stochastic
differential equations that
describe the diffusion process of the target
that is
to be learned.
The second feature is that it is based on a network that represents a
high-order weak approximation method for stochastic differential
equations.
The following sections provide an overview and background explanation
for each of these two features.
Henceforth, deep neural networks will be referred to as DNNs, neural
networks as NNs, and stochastic differential equations will be
abbreviated as SDEs.
%
\subsection{Deep neural network machines for learning diffusion processes}
DNN learning machines are a class of learning architectures that have
achieved remarkable success, beginning with pioneering work such as
\cite{4039068}, \cite{Fukushima1980NeocognitronAS}, and \cite{6795724}.
A DNN consists of numerous components, known as layers, arranged in a
serial configuration. Each layer receives a finite-dimensional vector
as input and produces a finite-dimensional output vector, i.e. it
implements a mapping from $\mathbb{R}^N$ to $\mathbb{R}^M$.
A common design involves composing a linear transformation from
$\mathbb{R}^N$ to $\mathbb{R}^M$ with a simple nonlinear function,
referred to as an activation function.
For simplicity, it is assumed in the following that both the input and output
dimensions of each layer are equal to $N$.
A learning machine composed of multiple such layers is referred to as a
multilayer neural network (MNN), and when the number of layers is large,
the architecture is termed a DNN.
\par
Among the many important discoveries in the progress of
DNN learning machines, the following three works are most
relevant to this present paper.
The first is~\cite{He-Zhang-Ren-Sun:7780459},
which proposes the so-called ResNet.
The second is~\cite{10.5555/3327757.3327764},
which proposes to regard the concept of ResNet as a numerical solver
of ordinary differential equations.
The third is~\cite{buehler2019deep}, which extends the proposal to the
case of SDEs and proposes a DNN learning machine to learn stochastic
processes. This method is successfully applied to hedging problems in
mathematical finance.

\subsection{High-order weak approximation algorithms for SDEs}
This section introduces two high-order discretisation methods for weak
approximation of SDEs, as proposed in
\cite{NinomiyaVictoir:2005} and
\cite{NinomiyaNinomiya:2007}.
These two methods have one notable feature in common.
This is that they can be implemented in the same way as the explicit
Runge--Kutta method,
i.e.  they can be implemented using only two types of operations:
linear combination of real vectors and function application.
\par
It is a natural approach to consider high-order discrete approximations
of SDEs using It\^o-Taylor expansions.
A series of studies on this topic, initiated by \cite{mil1975approximate},
has been widely referenced in numerous texts,
including \cite{KloedenPlaten:1999}.
However, the application and study of these approaches has been limited due to
the difficulty of dealing with iterative integrals of Brownian motion
in multidimensional cases, \mbox{i.e.} $d\geq 2$, in the notation below,
and the immaturity of high-dimensional numerical integration techniques.
\par
Quasi-Monte Carlo methods are subsequently found by
\cite{NinomiyaTezuka:1996} to be effective for high-dimensional numerical
integrals in the weak approximation of SDEs.
Then in \cite{kusuoka:2001aprx,kusuoka:2004revisited} and
\cite{LyonsVictoir:2002} these difficulties are overcome and the
theory of high-order discretisation of SDEs is established.
\subsubsection{Notation and conventions associated with SDEs}
Let $(\Omega,\mathcal{F},P)$ be a probability space,
$\begin{pmatrix}
  B^1(t) & \dots & B^d(t)
\end{pmatrix}$
be the $d$-dimensional standard Brownian motion,
$B^0(t)=t$,
and
$C^\infty_b(\mathbb{R}^N;\mathbb{R}^N)$
be the set of $\mathbb{R}^N$-valued smooth functions defined in $\mathbb{R}^N$
whose derivatives of any order are bounded.
$I_N$ denotes the identity
map on $\mathbb{R}^N$.
We also let
$X(t,x)$ denote a $\mathbb{R}^N$-valued
diffusion process defined by the
SDE~\cite{ikeda-watanabe}:
\begin{equation}\label{eq:SDE}
  X(t,x)=x+\sum_{i=0}^d\int_0^t V_iI_N(X(s,x))\circ dB^i(s).
\end{equation}
Where
$x\in\mathbb{R}^N$ and
$V_0,\dots,V_d$ are tangent vector fields on $\mathbb{R}^N$ whose coefficients
belong to $C^\infty_b(\mathbb{R}^N;\mathbb{R}^N)$,
\mbox{i.e.}
$V_iI_N =
\begin{pmatrix}
  (V_iI_N)^1 & \dots & (V_iI_N)^N
\end{pmatrix}
\in C_b^\infty(\mathbb{R}^N; \mathbb{R}^N)$
and
the equality
\begin{equation}\label{eq:vectorfieldnotation}
  V_ig(y)=\displaystyle{
    \sum_{j=1}^N (V_iI_N)^j(y)\frac{\partial g}{\partial x_j}(y)}
\end{equation}
holds for all $i\in\{0, 1, \dots, d\}$,
$g\in C^\infty(\mathbb{R}^N)$ and $y\in\mathbb{R}^N$.
Where $\circ\,dB^i(t)$ denotes the Stratonovich integral by $B^i(t)$.
Furthermore, SDE~\eqref{eq:SDE} can be expressed in an alternative form
by utilising the It\^o integral $dB^i(t)$:
\begin{equation}\label{eq:itoSDE}
  X(t,x)=x+\int_0^t \tilde{V_0}(X(s,x))\,dt
  +\sum_{i=1}^d\int_0^t V_i I_N(X(s,x))\, dB^i(s),
\end{equation}
where
\begin{equation}\label{eq:ito-stratonovich}
  (\tilde{V}_0I_N)^k=\left(V_0I_N\right)^k
  +\frac{1}{2}\sum_{i=1}^dV_i\left(V_iI_N\right)^k
\end{equation}
for all $k\in\{1,\dots,N\}$.
This equality is called It\^o--Stratonovich transformation.
\begin{rem}\label{rem:vecfield}
  Conventionally, a vector field $V$ and the vector composed of
  coefficient functions of $V$ are usually denoted by the same letter
  $V$.  However, in this paper, as we see in the discussion above, the
  latter is strictly denoted as $VI_N$.  This may seem redundant, but
  we dare to use this notation in this paper.
  This distinction contributes to the clear description of
  the It\^o--Stratonovich transformation~\eqref{eq:ito-stratonovich}
  introduced immediately above,
  as well as Definitions~\ref{dfn:resnet},
  \ref{def:rk5},
  \ref{dfn:NVnet}
  and \ref{dfn:NNnet}, which are to be seen in Section 2 below.
\end{rem}
\subsubsection{Weak approximation of SDEs}
Let $X(t,x)$ be a stochastic process defined by~\eqref{eq:SDE} and
$f$ be a $\mathbb{R}$-valued function defined on $\mathbb{R}^N$.
The numerical calculation of $E\left[f(X(T, x))\right]$ is referred to
as the weak approximation of SDE~\eqref{eq:SDE},
and it has been the focus of considerable research
due to its significance in practical applications
\cite{Glasserman:2004,KloedenPlaten:1999}.
\subsubsection{Simulation method}
Among the weak approximation methods for SDE,
the one relevant to this paper is the simulation method.
The weak approximation of SDE \eqref{eq:SDE}
by the simulation method is performed by the following procedure.
\par
Let $\Delta=\{0=t_0<t_1< \cdots <t_n=T\}$ be a partition of the interval
$[0,T]$.
As usual, we define
$\sharp\Delta = n$,
$\Delta_k=t_k-t_{k-1}$ for $k\in\{1,\dots,\sharp\Delta\}$
and
$
\lvert\Delta\rvert = \max\{t_{i+1}-t_i\,\vert\,0\leq i\leq \sharp\Delta-1\}.
$
We construct a set of random variables
$\left\{X^{(\Delta)}(t_i, x)\right\}_{i=0}^{\sharp\Delta}$
that approximates $\left\{X(t,x)\right\}_{0\leq t\leq T}$.
The pair of the partition $\Delta$ of $[0,T]$ and
this set of random variables
$\left(\Delta, \{X^{(\Delta)}(t_i, x)\}_{i=0}^{\sharp\Delta}\right)$
is called the discretisation of $X(t,x)$.
\par
It should be noted that
$X^{(\Delta)}(T,x)$
is an $\mathbb{R}^N$-valued function defined over a finite-dimensional domain.
If we denote the dimension of this domain by $D(\Delta)$, then we have a map
$X^{(\Delta)}(T,x)(\cdot):\mathbb{R}^{D(\Delta)}\to\mathbb{R}^N$.
Finally, the numerical integration $E\left[f(X^{(\Delta)}(T,x)\right]$
is performed.
This calculation is notorious as high-dimensional numerical integration
and we have to resort to Monte Carlo or quasi-Monte Carlo methods.
\subsubsection{Order of discretisation}\label{ss:order-of-discretisation}
As described above, the weak approximation calculation by the simulation
method is performed through two stages of discretisation and
numerical integration,  with approximation errors
occurring at each of these stages.
The approximation error generated in the former stage is referred to
as the discretisation error, while the error generated in the latter stage
is referred to as the integration error.
This paper deals only with the former type of error
and does not consider the latter.
This matter is discussed briefly in subsection~\ref{ss:qMC-MC-free} below.
\begin{dfn}\label{def:order-p}
  $\left\{X^{(\Delta)}(t_i, x)\right\}_{i=0}^{\sharp\Delta}$
  is defined to be a $p$th order discretisation
  or a discretisation of order $p$
  if there exists a positive $C_p$ and for all $n\in\mathbb{N}$ a partition
  $\Delta$ of $[0,T]$ such that $\sharp\Delta=n$ exists and the following
  inequality
  \begin{equation*}
    \left\lvert
    E[f(X(T,x))]-E[f(X^{(\Delta)}(T, x))]
    \right\rvert
    \leq C_p (\sharp\Delta)^{-p}
  \end{equation*}
  holds.
\end{dfn}
Note that this definition is made using $\sharp\Delta$, rather than $\lvert\Delta\rvert$. This is because we are concerned with the trade-off between computational accuracy and computational load.
\par
Henceforth, what is referred to as high-order discretisation in this paper
refers to discretisation of $2$nd order or higher.
\subsection{Discretisation methods}
We introduce some examples of discretisation methods.
For a tangent vector field $V$ on $\mathbb{R}^N$ and $x\in\mathbb{R}^N$,
$\exp(V)x$ denotes $z(1)$ where $z(t)$ is the solution
of the following ODE:
\begin{equation*}
  z(0)=x,\quad \frac{dz(t)}{dt}=VI_N(z(t)).
\end{equation*}
We remark that $\exp(tV)x=z(t)$ and
$(d/dt)(f(\exp(tV)x))=Vf(\exp(tV)x)$ hold.
We will refer to $\exp(V)(\exp(W)x)$ as $\exp(V)\circ \exp(W)x$.
\subsubsection{ODE solver, ODE integrator}
The algorithm or method for numerically computing $\exp(V)x$ is called
the ODE solver or ODE integrator.
\subsubsection{Euler--Maruyama
  \cite{Maruyama1955ContinuousMP,KloedenPlaten:1999}}
The Euler--Maruyama discretisation
$\left\{X^{(\rm{EM},\Delta)}(t_i,x)\right\}_{i=0}^{\sharp\Delta}$
is the most widely recognised technique for discretisation
of SDEs.
\begin{dfn}\label{def:EM}
  The Euler--Maruyama discretisation of SDE~\eqref{eq:SDE} is
  defined as follows:
  \begin{equation}\label{eq:EulerMaruyama}
    \begin{split}
      X^{(\rm{EM},\Delta)}(t_0,x) &= x \\
      X^{(\rm{EM},\Delta)}(t_{k},x)&= X^{(\rm{EM},\Delta)}(t_{k-1},x)
      +\Delta_{k}\tilde{V_0}I_N(X^{(\rm{EM},\Delta)}(t_{k-1},x)) \\
      &\quad+\sqrt{\Delta_{k}}
      \sum_{i=1}^dV_iI_N(X^{(\rm{EM},\Delta)}(t_{k-1},x))\eta^i_{k}
    \end{split}
  \end{equation}
  where $\left\{\eta^i_k\right\}_{\substack{1\leq i\leq d \\ 1\leq k\leq\sharp\Delta}}$
  is a family of independent random variables with the standard normal distribution.
\end{dfn}
For the Euler--Maruyama discretisation,
$D(\Delta) = d\times\sharp\Delta$.
Under certain conditions, the Euler--Maruyama discretisation achieves
$1$st order accuracy \cite{KloedenPlaten:1999}.
\subsubsection{Cubature 3 \cite{LyonsVictoir:2002,NinomiyaVictoir:2005}}
\begin{dfn}\label{def:cub3}
  Cubature~3 discretisation of SDE~\eqref{eq:SDE} is defined as follows:
  \begin{equation}\label{eq:cub3}
    \begin{split}
      X^{(\rm{cub3},\Delta)}(0,x) &= x \\
      X^{(\rm{cub3},\Delta)}(t_{k},x) &=
      \exp\left(
      \Delta_{k}V_0 +\sqrt{\Delta_{k}}\sum_{i=1}^d\eta_{k}^i V_i
      \right)X^{(\rm{cub3},\Delta)}(t_{k-1},x)
    \end{split}
  \end{equation}
  where $\left\{\eta^i_k\right\}_{\substack{1\leq i\leq d \\ 1\leq k\leq\sharp\Delta}}$
  is a family of independent random variables with the standard normal distribution.
\end{dfn}
According to \eqref{eq:cub3}, to find $X^{(\rm{cub3},\Delta)}(t_k,x)$,
one needs to start from $X^{(\rm{cub3},\Delta)}(t_{k-1},x)$ and proceed along
the vector field $\Delta_{k}V_0 +\sqrt{\Delta_{k}}\sum_{i=1}^d\eta_{k}^i V_i$
for time $1$.
\par
This method also achieves $1$st order accuracy and
$D(\Delta)=d\times\sharp\Delta$
\cite{NinomiyaVictoir:2005}.
\subsubsection{High-order method I \cite{NinomiyaVictoir:2005}}\label{ss:nv}
The method presented below is a high-order discretisation method for
weak approximation, possessing properties similar to those of the
explicit Runge--Kutta method. Specifically, it can be implemented
through iterations of function applications and linear combination
operations. Currently, only two high-order discretisation methods with
this property are known. These are the two methods described in this
subsection and the following subsection~\ref{ss:nn}
\begin{dfn}\label{def:nv}
  Let $\left\{\eta_k^i\right\}_{\substack{1\leq i\leq d \\ 1\leq k\leq\sharp\Delta}}$
  be
  a family of independent random variables with the standard normal distribution
  and $\left\{\Lambda_k\right\}_{1\leq k\leq\sharp\Delta}$ be a family of
  independent random variables such that
  $P(\Lambda_k=\pm 1)=1/2$ for all $k$.
  Then a discretisation
  $\{X^{(\rm{NV},\Delta)}(t_k,x)\}_{0\leq k\leq\sharp\Delta}$
  of the SDE~\eqref{eq:SDE}
  is defined as follows:
  \begin{equation}\begin{split}\label{eq:NV}
      & \,\, X^{(\rm{NV},\Delta)}(0,x) = x \\
      & X^{(\rm{NV},\Delta)}(t_{k},x) = \\
      &\qquad
      \begin{cases}
          &\exp\left(\displaystyle{\frac{\Delta_{k}}{2}}V_0\right)
          \left(
          \displaystyle{
            \prod_{i=1,\dots,d}^{\longrightarrow} \circ
            \exp\left(\sqrt{\Delta_{k}}\eta^i_{k}V_i\right)
          }
          \right)
          \circ\exp\left(\displaystyle{\frac{\Delta_{k}}{2}}V_0\right)
          X^{(\rm{NV},\Delta)}(t_{k-1},x) \\
          &\qquad\qquad\qquad\qquad\qquad\qquad\qquad\qquad\text{if $\Lambda_{k}=1$,}\\
          &\exp\left(\displaystyle{\frac{\Delta_{k}}{2}}V_0\right)
          \left(
          \displaystyle{
            \prod_{i=1,\dots,d}^{\longleftarrow} \circ
            \exp\left(\sqrt{\Delta_{k}}\eta^i_{k}V_i\right)
          }
          \right)
          \circ\exp\left(\displaystyle{\frac{\Delta_{k}}{2}}V_0\right)
          X^{(\rm{NV},\Delta)}(t_{k-1},x)\\
          &\qquad\qquad\qquad\qquad\qquad\qquad\qquad\qquad\text{if $\Lambda_{k}=-1$,}
      \end{cases}
  \end{split}\end{equation}
  where $\displaystyle{\prod_{i=1,\dots d}^{\longrightarrow}\circ A_i}
  =\circ A_1\circ A_2\circ\dots\circ A_d$ and
  $\displaystyle{\prod_{i=1,\dots d}^{\longleftarrow}\circ A_i}
  =\circ A_d\circ A_{d-1}\circ\dots\circ A_1$.
\end{dfn}
$\{X^{(\rm{NV},\Delta)}(t_k,x)\}_{0\leq k\leq\sharp\Delta}$
achieves $2$nd order discretisation
of SDE~\eqref{eq:SDE}
under certain conditions.
With a little ingenuity it is possible to set
$D(\Delta)=(d+1)\times\sharp\Delta$
in this method \cite{doi:10.1080/1350486X.2019.1637268}.
\par
As illustrated in Definition~\ref{def:cub3} and the subsequent discussion,
$X^{(\rm{NV},\Delta)}(t_k,x)$ is derived by solving ODEs with
$X^{(\rm{NV},\Delta)}(t_{k-1},x)$ as starting point.
The difference is that in this case, we have to solve $(d+2)$ ODEs
one by one by joining their solution curves.
\par
Let's describe this procedure in more detail.
First, the random variables $\Lambda_k$ and $\left\{\eta^i_k\right\}_{i=1}^d$
are drawn.
\par
In the case $\Lambda_k = 1$, starting from
$X^{(\rm{NV},\Delta)}(t_{k-1}, x)$, it moves along the vector field
$V_0$ for a time $\Delta_k / 2$. After this, it changes direction and
moves along the vector field $V_d$ for a time $\sqrt{\Delta_k}
\eta^d_k$. Then, it changes direction again and moves along the vector
field $V_{d-1}$ for a time $\sqrt{\Delta_k} \eta^{d-1}_k$. This motion
is repeated, and finally, after moving along the vector field $V_1$
for a time $\sqrt{\Delta_k} \eta^1_k$, it moves again along the vector
field $V_0$ for a time $\Delta_k / 2$, arriving at
$X^{(\rm{NV},\Delta)}(t_k, x)$. Note that time can be negative here.
\par
In the case $\Lambda_k = -1$, the procedure described above is
performed with the order of the vector fields reversed. That is,
starting from $X^{(\rm{NV},\Delta)}(t_{k-1}, x)$ and proceeding along
$V_0$ for a time $\Delta_k / 2$, then along the vector field $V_1$ for
a time $\sqrt{\Delta_k}\eta^1_k$, it changes direction and proceeds
along $V_2$, and finally along $V_d$, before proceeding along $V_0$
for a time $\Delta_k / 2$. The destination reached is
$X^{(\rm{NV},\Delta)}(t_k, x)$.
%
%
\subsubsection{High-order method II \cite{NinomiyaNinomiya:2007}}\label{ss:nn}
\begin{dfn}\label{def:nn}
  Let $\left\{\xi^i_k\right\}_{\substack{1\leq i\leq d \\ 1\leq k\leq\sharp\Delta}}$
  and $\left\{\eta^i_k\right\}_{\substack{1\leq i\leq d \\ 1\leq k\leq\sharp\Delta}}$
  be families of independent random variables with the
  standard normal distribution.
  Then a discretisation
  $\{X^{(\rm{NN},\Delta)}(t_k,x)\}_{0\leq k\leq\sharp\Delta}$
  of the SDE~\eqref{eq:SDE}
  is defined as follows:
  \begin{equation}\begin{split}
      X^{(\rm{NN},\Delta)}(0,x) &= 0 \\
      X^{(\rm{NN},\Delta)}(t_{k},x) &=
      \exp\left(
      c_1\Delta_{k}V_0+\sqrt{R_{11}\Delta_{k}}\sum_{i=1}^d\eta^i_{k}V_i
      \right) \\
      &\quad\circ
      \exp\left(
      c_2\Delta_{k}V_0+\sqrt{\Delta_{k}}\sum_{i=1}^d\zeta^i_{k}V_i      
      \right)X^{(\rm{NN},\Delta)}(t_{k-1},x)
  \end{split}\end{equation}
  where $u \geq 1/2$,
  \begin{gather*}
    c_1 = \mp\sqrt{(2u-1)/2},\quad c_2 = 1-c_1,\quad R_{11}=u, \\
    R_{22} = 1+u\pm\sqrt{2(2u-1)},\quad R_{12}= -u\mp\sqrt{(2u-1)/2}
  \end{gather*}
  and
  \begin{equation*}
    \zeta^i_{k}=\frac{R_{12}}{\sqrt{R_{11}}}\eta^i_k
    +\sqrt{R_{22}-\frac{{R_{12}}^2}{R_{11}}}\xi^i_k.
  \end{equation*}
\end{dfn}
$\{X^{(\rm{NN},\Delta)}(t_k,x)\}_{0\leq k\leq\sharp\Delta}$
also achieves $2$nd order discretisation
of SDE~\eqref{eq:SDE}
under certain conditions and
$D(\Delta)=2d\times\sharp\Delta$.
\subsubsection{Remark on cubature on Wiener space}
The subsequent paragraph offers a brief exposition of the relationships between the three approximation methods previously introduced --- Cubature 3 and the two high-order discretisation methods --- and the method of
cubature on Wiener space~\cite{LyonsVictoir:2002}.
\par
If we replace the standard normal random variables in
Definitions~\ref{def:cub3}, \ref{def:nv} and~\ref{def:nn}
with discrete random variables that are consistent with the normal distribution
up to a certain order of moments, we obtain cubatures on Wiener space.
In particular, by replacing the iid family
$\left\{\eta^i_k\right\}_{\substack{1\leq i\leq d \\ 1\leq k\leq\sharp\Delta}}$
in Definition~\ref{def:cub3} by such an iid family
$\left\{\hat{\eta}^i_k\right\}_{\substack{1\leq i\leq d \\ 1\leq k\leq\sharp\Delta}}$
defined as $P\left(\hat{\eta}^i_k=\pm 1\right)=1/2$
we obtain a cubature on Wiener space of order $1$.
Similarly, if we replace the iid families
$\left\{\eta^i_k\right\}_{\substack{1\leq i\leq d \\ 1\leq k\leq\sharp\Delta}}$
and
$\left\{\xi^i_k\right\}_{\substack{1\leq i\leq d \\ 1\leq k\leq\sharp\Delta}}$
in Definitions~\ref{def:nv} and \ref{def:nn}
by
such iid families
$\left\{\tilde{\eta}^i_k\right\}_{\substack{1\leq i\leq d \\ 1\leq k\leq\sharp\Delta}}$
and
$\left\{\tilde{\xi}^i_k\right\}_{\substack{1\leq i\leq d \\ 1\leq k\leq\sharp\Delta}}$
that are defined by
\begin{equation}\begin{split}
    \label{eq:5th-rv}
    P\left(\tilde{\eta}^i_k=\pm\sqrt{3}\right)
    =P\left(\tilde{\xi}^i_k=\pm\sqrt{3}\right) &=\frac{1}{6} \\
    P\left(\tilde{\eta}^i_k=0\right)
    =P\left(\tilde{\xi}^i_k=0\right)&=\frac{2}{3}
\end{split}\end{equation} respectively,
we obtain cubatures on Wiener space of order $2$.
For a more detailed explanation,
see \cite{LyonsVictoir:2002},
\cite{NinomiyaVictoir:2005} and \cite{NinomiyaNinomiya:2007}.
%
\section{High-order deep neural SDE network}\label{sec:highorderDNN}
\noindent This section introduces the concept of high-order deep neural SDE
networks.
\subsection{ResNet as Euler scheme}
The structure of ResNet can be regarded as an Euler approximation,
which is a typical numerical solution method for
ODEs~\cite{He-Zhang-Ren-Sun:7780459,10.5555/3327757.3327764}.
Let this first be explained below.
\subsubsection{Simple DNN and ResNet}
\begin{dfn}\label{dfn:resnet}
  A DNN is defined to be two finite sets of maps
  $\{F_k : \mathbb{R}^N\to\mathbb{R}^N\}_{k=1}^M$
  and $\{G_k : \mathbb{R}^N\to\mathbb{R}^N\}_{k=1}^M$.
  Each $F_k$ has a set of parameters $\{\alpha_i\}_{i=1}^{m_k}$
  which are to be learned through training.
  The $F_k$ is called as the $k$th layer
  and $M$ as the depth of the DNN.
  For maps $E$ and $F$, $E\circ F$ denotes the composition of them, \mbox{i.e.}
  $E\circ F(x) = E(F(x))$.
  $I_N$ denotes the identity map on $\mathbb{R}^N$.
  \begin{description}
  \item[Simple DNN]
    A DNN $\left\{\left(F_k, G_k\right)\right\}_{k=1}^M$
    is defined to be simple DNN when
    $G_1 = F_1$ and $G_k=F_k\circ G_{k-1}$ for $k\in\{2,\dots,M\}$.
  \item[ResNet]
    A DNN
    $\left\{\left(F_k, G_k\right)\right\}_{k=1}^M$ is defined to be ResNet when
    $G_1 = I_N + F_1$ and
    $G_k= G_{k-1} + F_k\circ G_{k-1}$ for $k\in\{2,\dots,M\}$.
  \end{description}
\end{dfn}
The term $G_{k-1}$ on the right hand side of the definition of ResNet
corresponds to the connection
called skipped connection or residual connection.
\subsubsection{Euler scheme and ResNet}\label{ss:eulerscheme}
We recall that
the Euler approximation
$\left\{x^{(\rm{Euler},\Delta)}(t_i,x_0)\right\}_{i=0}^{\sharp\Delta}$
with respect to a partition
$\Delta=\{0=t_0<t_1<\dots<t_{\sharp\Delta}=T\}$
of an $\mathbb{R}^N$-valued curve $\{x(t)\}_{t\in[0,T]}$
defined by the ODE:
\begin{equation}\label{eq:ODE}
  x(0)=x_0,\quad\frac{dx}{dt}(t)=VI_N(t, x(t)),
\end{equation}
which is equivalent to the alternative form
$x(t)=\exp(tV)x_0$
where $V$ is a tangent vector field on $\mathbb{R}^N$,
is defined as
\begin{equation}\begin{split}\label{eq:Euler}
    &x^{(\rm{Euler},\Delta)}(t_i,x_0)\\
    &=\begin{cases}
    x_0 &\text{if $i=0$} \\
    x^{(\rm{Euler},\Delta)}(t_{i-1},x_0)
    + \Delta_iVI_N \left(t_{i-1},x^{(\rm{Euler},\Delta)}(t_{i-1},x_0)\right) &\text{if $i\geq 1$}.
    \end{cases}
\end{split}\end{equation}
If we place $F_k$ in Definition~\ref{dfn:resnet}
as $F_k(\cdot) = \Delta_i VI_N(t_k,\cdot)$
it is easy to see that the correspondence:
\begin{equation}\label{eqn:euler-resnet}
  x^{(\rm{Euler},\Delta)}(t_k,x_0)=G_k(x_0)
\end{equation}
applies.
This means that the process of ResNet learning the parameter
$\alpha_k$ corresponds to learning the vector field $V$ at time $t_k$.
The Euler approximation is a first-order approximation.
With this in mind, we define ResNet as a first-order Net.
\subsection{High-order approximation scheme for ODEs and neural net}
In light of the observations made in subsection~\ref{ss:eulerscheme}
regarding first-order approximation methods, we proceed to construct
neural networks corresponding to high-order approximation methods. It
is notable that the neural network corresponding to the explicit
approximation method of the Runge--Kutta type is relatively
straightforward to construct.
\subsubsection{$5$th order neural net}\label{ss:RK5}
From \cite{butcher:1987} we introduce an explicit $5$th order Runge--Kutta type
approximation.
\begin{dfn}\label{def:RK5}
  We define an approximation
  $\{x^{(\rm{RK5},\Delta)}(t_i,x_0)\}_{i=0}^{\sharp\Delta}$
  of the ODE~\eqref{eq:ODE} with respect to $\Delta$
  as follows:
  \begin{equation}\label{eq:RK5}
    \begin{split}
      \text{For $i\in\{1,2,\dots,6\}$}\quad
      Y_i &= x^{(\rm{RK5},\Delta)}(t_{k-1},x_0)+\Delta_k\sum_{j<i}a_{ij}Z_j \\
      Z_i &=VI_N(Y_i) \\
      x^{(\rm{RK5},\Delta)}(t_k, x_0) &=x^{(\rm{RK5},\Delta)}(t_{k-1},x_0)+\Delta_k\sum_{j=1}^6b_jZ_j
    \end{split}
  \end{equation}
  where
  \begin{gather*}
    a_{21}=2/5,\quad a_{31}=11/64,\quad a_{32}=5/64,\quad a_{43}=1/2,
    \quad a_{51}=3/64, \\
    a_{52}=-15/64,\quad a_{53}=3/8,\quad a_{54}=9/16,\quad a_{62}=5/7,
    \quad a_{63}=6/7, \\
    a_{64}= -12/7,\quad a_{65}=8/7,\\
    a_{ij}=0\quad\text{otherwise},\\
    b=\begin{pmatrix}
    \displaystyle{\frac{7}{90}} & 0 &
    \displaystyle{\frac{32}{90}} & \displaystyle{\frac{12}{90}} &
    \displaystyle{\frac{32}{90}} & \displaystyle{\frac{7}{90}}
    \end{pmatrix}.
  \end{gather*}
  This method
  achieves a $5$th order approximation.
\end{dfn}
From this definition, it can be seen that this approximation method
has a similar form to the explicit Runge--Kutta type. That is, the
method can be performed by repeating only two types of operations: the
application of functions and the linear combination of the
results.
Furthermore, this sequence of operations does not include
forward references.
These features allow the method to be implemented
as a neural network, as the definition and subsequent Figure~\ref{fig:rk5}
below show.
\begin{dfn}[RK5net]\label{def:rk5}
  A DNN
  $\left\{(F_k, G_k)\right\}_{k=0}^{\sharp\Delta}$
  is defined to be RK5net as 
  $F_k(\cdot)=VI_N(t_k,\cdot)$,
  \begin{equation}
    \begin{split}
      H_1 & = I_N \\
      H_2 &= I_N+\Delta_k a_{21}F_k \\
      H_3 &= I_N+\Delta_k\left(a_{32}F_k\circ H_2+(a_{31}+a_{21}\right)F_k) \\
      H_4 &= I_N+\Delta_k\left(a_{43}F_k\circ H_3+a_{32}F_k\circ H_2+(a_{31}+a_{21}\right)F_k) \\
      H_5 &= I_N+\Delta_k\left(a_{54}F_k\circ H_4+(a_{53}+a_{43})F_k\circ H_3
      +(a_{52}+a_{32})F_k\circ H_2\right. \\
      &\quad+\left.(a_{51}+a_{31}+a_{21})F_k\right) \\
      H_6 &= I_N+\Delta_k\left(a_{65}F_k\circ H_5+(a_{64}+a_{54})F_k\circ H_4
      +(a_{63}+a_{53}+a_{43})F_k\circ H_3\right. \\
      &\quad+\left.(a_{62}+a_{52}+a_{32})F_k\circ H_2
      +(a_{51}+a_{31}+a_{21})F_k,
      \right)
    \end{split}
  \end{equation}
  and
  \begin{equation}
    G_k = \left(I_N + \Delta_k\sum_{j=1}^6 H_j\right)\circ G_{k-1},
  \end{equation}
  where constants $\left\{a_{ij}\right\}_{1\leq i,j\leq 6}$ and $\left\{b_i\right\}_{j=1}^6$
  are the same as those defined in Definition~\ref{def:RK5}.
\end{dfn}

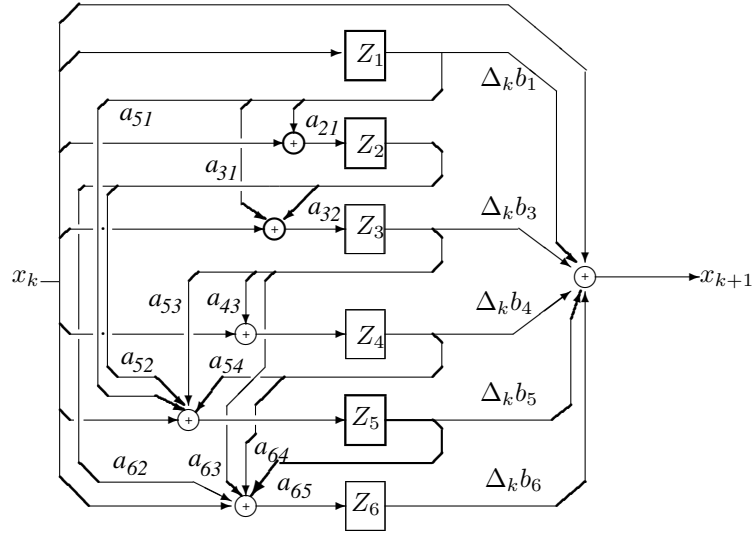
\begin{figure}[h]
  \input{RK_5.latex}
  \captionof{figure}{$5$th order explicit Runge--Kutta type method}
  \label{fig:rk5}
\end{figure}
%
\subsubsection{Orders of weak approximations for SDEs and
  those of approximation methods for ODEs}
The weak approximation methods of SDEs introduced in
Definitions~\ref{def:cub3},~\ref{def:nv} and ~\ref{def:nn}
all entail integral of some ODEs, \mbox{i.e.} numerical calculation of $\exp(tV)$,
as part of their implementation.
The following theorem, which concerns the relationship between the
order of the weak approximation method for SDEs and the order of
the approximation method for ODEs contained therein, is
guaranteed to hold by Theorem~\mbox{1.3} in~\cite{NinomiyaNinomiya:2007}.
\begin{thm}\label{thm:order}
  The discretisation method for SDE~\eqref{eq:SDE},
  as defined in Definition~\ref{def:cub3},
  preserves order $1$ if the ODEs it contains are solved using an ODE solver
  of order $3$ or higher.
  Similarly, the methods defined in Definitions~\ref{def:nv} and \ref{def:nn}
  preserve order $2$ if the ODEs they contain are solved using an ODE solver
  of order $5$ or higher.
\end{thm}
The claim of this theorem holds for more general $p$th order discrete
approximation methods, including the three discretisation methods
mentioned in the theorem, but to state the claim it is necessary to
define the concept of an ODE-valued random variable, which is omitted
here. See Theorem~\mbox{1.3} in~\cite{NinomiyaNinomiya:2007} for the
exact claim.
\subsection{High-order deep neural SDE network}
The concept of a deep neural SDE network, which is a neural network
that learns stochastic differential equations, is initially proposed
in \cite{buehler2019deep}.  In this paper, a first-order method
analogous to ResNet is utilised for the discretisation of SDEs.
\par
In this section, we build on the preparation of the previous sections
and put forward the concept of a high-order deep neural SDE network.
\subsubsection{High-order deep neural SDE network: \rm{NVnet}}
$\{X^{(\rm{NV},\Delta)}(t_k,x)\}_{k=0}^{\sharp\Delta}$ in Definition~\ref{def:nv}
is implemented by series-connecting
$(d+2)$ ODE integrators. If each of these ODE integrator is
implemented with the RK5net given in Definition~\ref{def:RK5},
a neural network
realising $\{X^{(\rm{NV},\Delta)}(t_k,x)\}_{k=0}^{\sharp\Delta}$ is obtained.
\begin{dfn}[NVnet]\label{dfn:NVnet}
  Let
  $\left\{\eta_k^i\right\}_{\substack{1\leq i\leq d \\ 1\leq k\leq\sharp\Delta}}$
  be a family of independent random variables with the
  standard normal distribution
  and $\left\{\Lambda_k\right\}_{1\leq k\leq\sharp\Delta}$ be a family of
  independent random variables such that
  $P(\Lambda_k=\pm 1)=1/2$ for all $k$.
  NVnet $\left\{\left(G_k, F_k\right)\right\}_{k=1}^{\sharp\Delta}$
  is a DNN defined as:
  \begin{itemize}
  \item $F_k$, the $k$th layer of NVnet is composed of
    $(d+2)$ ODE integrators, which are connected in series
    as follows:
    \begin{equation}\begin{split}
        &\exp\left(\displaystyle{\frac{\Delta_{k}}{2}}V_0(t_k,\cdot)\right)
        \left(
        \displaystyle{
          \prod_{i=1,\dots,d}^{\longrightarrow} \circ
          \exp\left(\sqrt{\Delta_{k}}\eta^i_{k}V_i(t_k,\cdot)\right)
        }
        \right)
        \circ
        \exp\left(\displaystyle{\frac{\Delta_{k}}{2}}V_0(t_k,\cdot)\right)\\
        &\qquad\qquad\qquad\qquad
        \qquad\qquad\qquad\qquad\qquad\qquad\qquad\qquad\qquad
        \text{if $\Lambda_k=1$,}
        \\
        &\exp\left(\displaystyle{\frac{\Delta_{k}}{2}}V_0(t_k,\cdot)\right)
        \left(
        \displaystyle{
          \prod_{i=1,\dots,d}^{\longleftarrow} \circ
          \exp\left(\sqrt{\Delta_{k}}\eta^i_{k}V_i(t_k,\cdot)\right)
        }
        \right)
        \circ
        \exp\left(\displaystyle{\frac{\Delta_{k}}{2}}V_0(t_k,\cdot)\right)\\
        &\qquad\qquad\qquad\qquad
        \qquad\qquad\qquad\qquad\qquad\qquad\qquad\qquad\qquad
        \text{if $\Lambda_k=-1$}.
    \end{split}\end{equation}
  \item All ODE integrators involved are implemented using RK5net.
  \item All $\sharp\Delta$ layers are connected in series,
    i.e.
    $G_1=F_1$ and $G_k=F_k\circ G_{k-1}$ for $2\leq k\leq\sharp\Delta$.
  \end{itemize}
\end{dfn}
It follows from Theorem~\ref{thm:order} that NVnet is
a $2$nd order neural SDE network. This network learns the vector fields
$V_i$, which is equivalent to learning the stochastic process
described by SDE~\eqref{eq:SDE}.
\subsubsection{High-order deep neural SDE network: \rm{NNnet}}
$\{X^{(\rm{NN},\Delta)}(t_k,x)\}_{0\leq k\leq\sharp\Delta}$
is implemented by series connection of two ODE integrators
from Definition~\ref{def:nn}.
Thus, as in the case of NVnet, a neural network implementation of this
can be obtained by using RK5net as an aid.
\begin{dfn}[NNnet]\label{dfn:NNnet}
  Let $\left\{\xi^i_k\right\}_{\substack{1\leq i\leq d \\ 1\leq k\leq\sharp\Delta}}$,
  $\left\{\eta^i_k\right\}_{\substack{1\leq i\leq d \\ 1\leq k\leq\sharp\Delta}}$,
  $u$,  $c_1$, $c_2$, $R_{11}$, $R_{12}$, $R_{22}$ and $\zeta^i_k$ be
  the same as previously defined in Definition~\ref{def:nn}.
  NNnet is a deep neural network
  $\left\{\left(F_k, G_k\right)\right\}_{k=1}^{\sharp\Delta}$
  defined as:
  \begin{itemize}
  \item $F_k$, the $k$th layer of NNnet
    is composed of two ODE integrators,
    which are connected in series as follows:
    \begin{equation}\begin{split}
      &\exp\left(
      c_1\Delta_{k}V_0+\sqrt{R_{11}\Delta_{k}}\sum_{i=1}^d\eta^i_{k}V_i(t_k,\cdot)
      \right)\\
      &\qquad\qquad\circ
      \exp\left(
      c_2\Delta_{k}V_0+\sqrt{\Delta_{k}}\sum_{i=1}^d\zeta^i_{k}V_i(t_k,\cdot)
      \right).
    \end{split}\end{equation}
  \item All ODE integrators involved are implemented using RK5net.
  \item All $\sharp\Delta$ layers are connected in series,
    i.e. 
    $G_1=F_1$ and
    $G_k=F_k\circ G_{k-1}$ for $2\leq k\leq \sharp\Delta$.
  \end{itemize}
\end{dfn}
Similarly to NVnet, NNnet is a $2$nd order deep neural SDE
network, for the same reasons.
The vector fields $V_i$s are to be learnt.
\section{Numerical experiment}
\noindent
The calculations are conducted using two types of deep neural SDE
networks: NVnet, which is proposed in the preceding section, and ResNet
for comparison.
The networks are employed to compute the hedge martingale for an
at-the-money (ATM) American vanilla put option.
\subsection{American option pricing by deep neural SDE network}
We let $\left(\Omega,\mathcal{F}, P\right)$
be a probability space with Brownian filtration
$(\mathcal{F}_t)_{t\in [0,T]}$,
$\mathcal{S}_T$ be the set of
all $(\mathcal{F}_t)_{t\in [0,T]}$-stopping times bounded by $T$
and $(Z_t)_{t\in [0,T]}$ be the payoff process of some American option.
It is widely acknowledged
\cite{10.1093/oso/9780198851615.001.0001,Glasserman:2004}
that the fair price of this American option is given by
$
\displaystyle{\sup_{\tau\in\mathcal{S}_T} E\left[Z_\tau\right]}
$
and that it is notoriously challenging
to calculate prices using
the simulation method according to this formula. This difficulty
arises from the fact that the formula involves optimisation over all
stopping times bounded by $T$.
\subsubsection{Rogers' dual algorithm}
In \cite{https://doi.org/10.1111/1467-9965.02010}, an expression free
from stopping times for the appropriate price of the American option
is given in accordance with the following theorem.
\begin{thm}\label{thm:rogers}
  The price of an American option whose payoff process is
  given by
  $(Z_t)_{t\in[0,T]}$ is equal to
  \begin{equation}\label{eq:AOprice}
    \inf_{M\in H^1_0}E\left[\sup_{t\in[0,T]}(Z_t-M_t)\right]
  \end{equation}
  where
  \begin{equation*}
    H^1_0=\left\{M\,\left\vert\,
    \text{$(\mathcal{F}_t)_{t\in[0,T]}${\rm -martingale s.t.}
      $\displaystyle{\sup_{t\in[0,T]}\lvert M_t\rvert\in L^1(P)}$, $M_0=0$}
    \right.\right\}.
  \end{equation*}
\end{thm}
This expression does not explore the space of stopping
times; instead, it explores the space for martingales and identifies
the lower bound of the maximum process of the difference between the
payoff and the martingale. This is essentially a search for the
optimal hedging process for the target American option.
\subsubsection{Pricing process by learning machines}\label{ss:slogan}
According to Theorem~\ref{thm:rogers},
the determination of the price of an American
option by a learning machine amounts to the following slogan:
\begin{equation}\label{slogan}
  \text{Find $M^*\in\displaystyle{\arginf_{M\in H^1_0}E\left[
        \sup_{t\in[0,T]}(Z_t-M_t)
        \right]}$ by learning.}
\end{equation}
This is carried out in two steps:
\begin{description}
\item[Step 1] Express the martingale $M_t\in H^1_0$ as
  \begin{equation*}
    M_t=\int_0^t V_0^MI_N\left((X_u, M_u)_{0\leq u\leq s}\right)\,ds
    +\sum_{j=1}^d\int_0^t V_j^MI_N\left((X_u, M_u)_{0\leq u\leq s}\right)
    \circ\,dB^j(s)
  \end{equation*}
  implementing functions $V_0^MI_N, V_1^MI_N, \dots, V_d^MI_N$
  in the form of simple MLP(=Multi Layer Perceptron)~\cite{Geron2019book}.
\item[Step 2]
  Optimise all $V_j^MI_n$s to minimise
  $E\left[\displaystyle{\sup_{t\in[0,T]}(Z_t-M_t)}\right]$ by using
  deep neural SDE network.
\end{description}

\subsection{Important details of the implementation}

This section discusses two technical issues that significantly
influence the implementation of the proposed method. The first
concerns the difficulty of incorporating the It\^o--Stratonovich
transformation~\eqref{eq:ito-stratonovich} into a network architecture
composed of MLPs. The second involves the challenge of obtaining an
accurate maximum of a stochastic process under high-order
discretisation.

\subsubsection{Obstacle arising from the It\^o--Stratonovich transformation}

As noted in subsection~\ref{ss:slogan}, the vector fields $V_0^M,
V_1^M, \dots, V_d^M$ are modelled using MLPs, whose parameters are
learned during training. A central concern is ensuring that the
resulting process $M_t$ is a martingale. This concern arises because
the networks (e.g., NVnet and NNnet) approximate the Stratonovich
formulation of the dynamics, and hence, naively omitting the $dt$-term
does not guarantee the martingale property.
\par
To ensure correctness, one would need to realise the
It\^o--Stratonovich transformation within the neural
network. Specifically, the transformation~\eqref{eq:ito-stratonovich},
\begin{equation*}
  \left(V^M_0I_N\right)^k
  +\frac{1}{2}\sum_{i=1}^dV^M_i\left(V_i^MI_N\right)^k,
\end{equation*}
requires evaluating derivatives of vector fields (represented by MLPs)
applied to their own coefficient functions, which are represented by
the same MLPs. While it may be theoretically possible to construct a
network that expresses this computation, it raises interesting
challenges that fall outside the scope of this paper.
\subsubsection{A low-cost surrogate for the canonical method}
\label{ss:MeasureNeutral}
A practical workaround to the above issue is available and avoids the
aforementioned difficulties.
Although the solution may appear
counterintuitive, it proves effective in practice.
The key idea is to design the implementation so that the process $M_t$
remains centred across all sample paths, thereby satisfying the least
martingale condition in a distributional sense, even though $M_t$ may
not fully satisfy all properties of a true martingale.
\par
This is achieved as follows:
\begin{description}
\item[Step~1] Exclude $V^M_0$ by setting $V^M_0 = 0$.
\item[Step~2] For each path $\omega_i$, generate a provisional path
  $M_t'(\omega_i)$ using the network defined by the remaining vector
  fields $V_1^M, \dots, V_d^M$.
\item[Step~3] Define the actual process $M_t(\omega_i)$ as a centred version:
  \begin{equation}\label{eq:MeasureNeutral}
    M_t(\omega_i) = M_t'(\omega_i) - \frac{1}{K_{\rm{BIN}}}
    \sum_{j=1}^{K_{\rm{BIN}}}M_t'(\omega_j),
  \end{equation}
  where $K_{\rm{BIN}}$ denotes the batch size used to estimate the expectation.
\item[Step~4] Use the centred process $M_t$ to evaluate the objective
  function in Rogers' dual formulation:
  \begin{equation*}
    \frac{1}{K_{\rm{BIN}}}\sum_{i=1}^{K_{\rm{BIN}}}
    \sup_{t\in [0,T]}\left(Z_t(\omega_i)-M_t(\omega_i)\right).
  \end{equation*}
\end{description}
This centring operation removes the drift component that would
otherwise arise due to the absence of the $V^M_0$ term.
\subsubsection{A reference implementation of the canonical method}
To provide context we briefly describe a canonical method that
directly enforces the martingale property at the cost of a
significantly increased computational workload.
\par
In the canonical method, a path
$\left\{M_{t_i}(\omega)\right\}_{i=0}^{\sharp\Delta}$
is constructed as follows:
\begin{description}
\item[Step~1] Set $M_0(\omega) = 0$.
\item[Step~2] For each $i\in\{1\dots\sharp\Delta\}$,
  draw $K$ independent
  realisations of $M^\prime_{t_i}$ under the condition that
  $M^\prime_{t_{i-1}} = M_{t_{i-1}}(\omega)$.
  We denote this set by
  $\left\{M^\prime_{i,j}(\omega)\right\}_{j=1}^{K}$.
  Then define
  \begin{equation*}
    M_{t_i}(\omega)
    = M^\prime_{i,1}(\omega)
    - \frac{1}{K}\sum_{j=1}^{K} M^\prime_{i,j}(\omega).
  \end{equation*}
\end{description}
Here, $K$ is a sufficiently large positive integer
and $\Delta = (0 = t_0 < t_1 < \dots < t_{\sharp\Delta} = T)$ denotes
a partition of the interval $[0,T]$.
The component $V^M_0$ is set to some value, say $0$.
Each $M^\prime_{t_i}(\omega)$ is generated using the network defined by
the vector fields $V^M_1, \dots, V^M_d$.
\par
It is readily seen that the computational cost of this procedure is at least
$K$ times greater than that of the low-cost
version.
This leaves the surrogate method as the viable option for practical use.
\subsubsection{Simulation of the maximum process}
It is well known that when trying to find
$\sup_{t\in [0,T]}Y_t(\omega)$
by simulation with respect to a partition
$\Delta$, the naive method of taking
$\max\{Y_i\,\vert\,i\in\{0,1,\dots,\sharp\Delta\}\}$ as the value,
is subject to such large errors that it becomes almost unusable as the
width of the partition increases.
This issue cannot be overlooked, particularly in our context. The
NVnet and NNnet are founded upon high-order discrete approximations,
which offer the primary benefit of enabling the partition width to be
significantly expanded.
\par
The well-known method of approximating
$\sup_{t\in[t_{k-1},t_k]}(Z_t(\omega)-M_t(\omega))$ using the Brownian
bridge overcomes this difficulty.
We regard $\displaystyle{\sup_{t\in [t_{k-1},t_k]}(Z_t(\omega)-M_t(\omega))}$
as
\begin{equation}\label{eq:BB}
  \sup_{t\in [t_{k-1},t_k]}\left\{\sigma B_t\,\left\vert\,
  B_{t_{k-1}}=\frac{a}{\sigma}\quad\text{and}\quad
  B_{t_k}=\frac{b}{\sigma}\right.\right\}
\end{equation}
where $a=Z_{t_{k-1}}(\omega)-M_{t_{k-1}}(\omega)$ and
$b=Z_{t_k}(\omega)-M_{t_k}(\omega)$.
It is then necessary to find the value of the volatility $\sigma$.  We
draw about $10$ sample paths from $Z_{t_{k-1}}-M_{t_{k-1}}$ and
adopted their sample volatility.
According to the well-known distribution function of pinned
Brownian motion \cite{ikeda-watanabe},
we can calculate the cumulative density function
$F_{\sigma B}$ of the pinned Brownian motion~\eqref{eq:BB}
and its inverse function $G_{\sigma B}$ as follows:
\begin{equation}\begin{split}
    F_{\sigma B}(x) &= \int_{a\vee b}^x\frac{2(2y-b-a)}{\Delta_k}
    \exp\left(
    \frac{2(a-y)(y-b)}{\Delta_k}
    \right)\,dy \\
    G_{\sigma B}(p) &= \frac{1}{2}\left(
    a+b+\sqrt{(a-b)^2-2\sigma^2\Delta_k\log(1-p)}\right).
\end{split}\end{equation}
These can be employed to perform a simulation with a minimal
computational load.
\subsection{Specifications and parameters}

\subsubsection{Computer and software used}
Numerical experiments are conducted on a computational system equipped
with an Intel i9-13900K CPU and an NVIDIA GTX4090 GPU.
All numerical experiment in this article was programmed with TensorFlow~2,
and parallel computation on GPU is employed.
\subsubsection{Common parameters}\label{ss:CommonParameters}
The following specifications and parameters are commonly used in all numerical calculations in this study.
\par
Each vector field $V_j^M$s is implemented by an MLP \cite{Geron2019book}
consisting of two hidden
layers, each with 32 nodes, and an output layer, also with 32 nodes.
The ReLU activation function~\cite{Geron2019book}
is applied across all layers to ensure
non-linearity and effective learning.
\par
All numerical integrations included are proceeded by using Quasi-Monte
Carlo method with $K_{\rm BIN}=5000$ sample points generated
from the generalised
Sobol' sequence~\cite{NinomiyaTezuka:1996,Glasserman:2004}.
As described below (in subsection~\ref{ss:ProcCycle}),
the parameters are updated at each evaluation in this experiment, i.e. EPOC=1.
\par
The result of Theorem~\ref{thm:rogers} allows for the utilisation
of the value of~\eqref{eq:AOprice} itself as the LOSS
function~\cite{Geron2019book} for this learning.
\subsubsection{Algorithm for updating parameters}
The Adam optimiser~\cite{Kingma2014AdamAM, Geron2019book},
which is provided by the TensorFlow library, is utilised
with learning\_rate=0.001, beta\_1=0.9, beta\_2=0.999, epsilon=1e-07 and
2000 learning iterations are conducted to train MLPs.
\par
Adam optimiser is an optimiser that updates parameters
via dynamically scaling the update step based on moment estimation
along each parameter's gradient vector.

Assuming we have a $\mathrm{LOSS}$ function which parameters
are $\theta$. Within Adam optimiser, we expect $\theta$ is updated
according to its gradient w.r.t. $\mathrm{LOSS}$ function. Here,
$\theta$ is the all parameters in MLPs.
Adam optimiser differs from standard Stochastic Gradient Descent (SGD),
Newton-type methods, and other non-parametric optimisers.
SGD uses sorely
gradient information, and Newton's method uses gradient and hessian
information, however, they do not employ statistical moment to accelerate
their parameter updating. Nevertheless, Adam optimiser takes advantage
of $1^{\mathrm{st}}$ and $2^{\mathrm{nd}}$ statistical moment estimation.
In the Adam optimiser, the moment can be regarded simply as weighted
expectation of the mean and the variance of gradients. Since Adam
optimiser avoid burdensome and memory leaking Hessian computation,
this feature results Adam optimiser in a faster and well-behaved
performance when it engages large scaled optimisation problems. 

We briefly exhibit mechanism of Adam optimiser below: 

STEP1: $1^{\mathrm{st}}$ and $2^{\mathrm{nd}}$ Moment Initialisation:
\[
\mathbf{m}_{0}=0,\qquad\mathbf{v}_{0}=0,\qquad t=0
\]

STEP2: Gradient Computation:

\[
\mathbf{g}_{t}=\nabla_{\mathbf{\theta}}\mathrm{LOSS}\left(\mathbf{\theta}_{t-1}\right)
\]

STEP3: $1^{\mathrm{st}}$ and $2^{\mathrm{nd}}$ Moment Estimation:

\[
\mathbf{m}_{t}=\beta_{1}\mathbf{m}_{t-1}+(1-\beta_{1})\mathbf{g}_{t}
\]
\[
\mathbf{v}_{t}=\beta_{2}\mathbf{v}_{t-1}+(1-\beta_{2})\mathbf{g}_{t}\circ\mathbf{g}_{t}
\]

STEP4: Bias Correction on $1^{\mathrm{st}}$ and $2^{\mathrm{nd}}$
Moment:

\[
\hat{\mathbf{m}_{t}}=\frac{\mathbf{m}_{t}}{1-\beta_{1}^{t}}
\]

\[
\hat{\mathbf{v}_{t}}=\frac{\mathbf{v}_{t}}{1-\beta_{1}^{t}}
\]

STEP5: Parameter Updating:

\[
\mathbf{\theta}_{t}=\mathbf{\theta}_{t-1}-\alpha\frac{\hat{\mathbf{m}_{t}}}{\sqrt{\hat{\mathbf{v}_{t}}}+\epsilon}
\]

where $\alpha$ is the learning rate, $\circ$ is Hadamard production
and the term
$\frac{\hat{\mathbf{m}_{t}}}{\sqrt{\hat{\mathbf{v}_{t}}}+\epsilon}$
is computed as element-wise.
\par
All numerical integrations included are proceeded by using Quasi-Monte
Carlo method with $K_{{\rm BIN}}=5000$ sample points generated from
the generalised Sobol' sequence~\cite{NinomiyaTezuka:1996,Glasserman:2004}.
As described below (in subsection~\ref{ss:ProcCycle}), the parameters
are updated at each evaluation in this experiment, i.e. EPOC=1. 
\par
The result of Theorem~\ref{thm:rogers} allows for the utilisation
of the value of~\eqref{eq:AOprice} itself as the
LOSS function~\cite{Geron2019book} for this learning. 
\subsubsection{Asset price models}
The calculations are conducted for each case in which the risky asset
follows two models: the Black--Scholes--Merton model~\cite{BlackScholes:1973}
and the Heston stochastic volatility model~\cite{heston:1993}.
\par
The Black--Scholes--Merton model is obtained by setting
$N=1$,
$X_t=S_t$,
$V_0I_1(y)=\left(\mu-\sigma^2/2\right)y$
and
$V_1I_1(y)=\sigma y$ in SDE~\eqref{eq:SDE}.
In the calculations with this model, we set $S_0=100$,
$\mu=0$
and $\sigma=0.32$.
\par
Furthermore, SDE~\eqref{eq:SDE} with
$N=2$,
$X_t=\prescript{t}{}{
  \begin{pmatrix}
    S_t & U_t
  \end{pmatrix}
}$,
\begin{equation*}\begin{split}
    V_0I_2\left(
    \prescript{t}{}{\begin{pmatrix} y_1 & y_2 \end{pmatrix}}
    \right)
    &=\prescript{t}{}{
      \begin{pmatrix}
        \left(\mu-y_2/2-\rho\beta/4\right)y_1 &
        \alpha(\theta - y_2)-\beta^2/4
      \end{pmatrix}
    },
    \\
    V_1I_2\left(
    \prescript{t}{}{\begin{pmatrix}y_1 & y_2\end{pmatrix}}
    \right)
    &=\prescript{t}{}{
      \begin{pmatrix}
        y_1\sqrt{y_2} & \rho\beta\sqrt{y_2}
      \end{pmatrix}
    }
    \\
    \text{and}\quad
    V_2I_2\left(
    \prescript{t}{}{\begin{pmatrix}y_1 & y_2\end{pmatrix}}
    \right)
    &=\prescript{t}{}{
      \begin{pmatrix}
        0 & \beta\sqrt{(1-\rho^2)y_2}
      \end{pmatrix}
    }
  \end{split}
\end{equation*}
yields the Heston stochastic volatility model.
Here, $S_t$ represents the asset price at time $t$ and
$U_t$ denotes its variance at that time.
In the calculations with this model, we set
$S_0=100$, $U_0=0.32$,
$\mu=0$,
$\theta=0.25$,
$\alpha=3.0$,
$\rho = 0.3$
and
$\beta=0.4$.
\par
Our target is an American put option on the asset whose price at $t$ is $S_t$
with a strike price of $K=100$ and expiration date of $T=1.0$.
The payoff process $Z_t$ is expressed as $Z_t=\max\{K-S_t, 0\}$.
\subsubsection{Implementation of $M_t$}
As described in subsection~\ref{ss:slogan},
each vector fields of $M_t$ is implemented by an MLP.
In this experiment, the MLPs are constrained to be of the form having
$(t,X_t,M_t)$ as input and output.
In general, this restriction makes the space of martingales
in which $M_t$ moves smaller than $H^1_0$.
However, we ignore it here.
\subsubsection{Procedure for one learning cycle}\label{ss:ProcCycle}
The following procedure is to be employed each time
the MLP parameters are updated.
\begin{description}
\item[Step 1] Draw
  $\left\{
  \{X_{t_j}^{(\rm{Alg},\Delta)}(\omega_i)\}_{j=0}^{\sharp\Delta}
  \right\}_{i=1}^{K_{\rm BIN}}$
  and
  $\left\{\{M_{t_j}(\omega_i)\}_{j=0}^{\sharp\Delta}\right\}_{i=1}^{K_{\rm BIN}}$.
  The former is generated in accordance with equation~\eqref{eq:EulerMaruyama}
  when {\bf Alg} is {\bf EM} and in accordance with equation~\eqref{eq:NV}
  when {\bf Alg} is {\bf NV}.
  The latter is generated in accordance with equation~\eqref{eq:MeasureNeutral}
  by ResNet if {\bf Alg} is {\bf EM} or by NVnet if {\bf Alg} is {\bf NV},
  following
  the procedure outlined in subsection~\ref{ss:MeasureNeutral}.
\item[Step 2]
  Approximate
  $E\left[\displaystyle{\sup_{t\in[0,T]}(Z_t-M_t)}\right]$
  by the sample mean
  \begin{equation}\label{eq:SampleMean}
    \frac{1}{K_{\rm BIN}}
    \sum_{i=1}^{K_{\rm BIN}}
    \sup_{t\in [0,T]}\left(
    Z_t(\omega_i)-M_t(\omega_i)
    \right)
  \end{equation}
  which is calculated from
  the samples drawn in the previous step.
  Subsequently, the parameters of the MLP are updated in order to reduce
  the value of the expression given by \eqref{eq:SampleMean}.
\end{description}
\subsection{Results}\label{ss:results}
Figures~\ref{fig:BS} and \ref{fig:HESTON}
show the results of the numerical experiments.
In both figures, a comparison is made between learning with ResNet
and learning with NVnet, one of the new architectures proposed in this paper.
The vertical axis represents the LOSS and the
horizontal axis represents the number of parameter updates,
\mbox{i.e.} the number of learning iterations.
Figure~\ref{fig:BS} shows the case where asset price follows the
Black--Scholes--Merton model, while Figure~\ref{fig:HESTON} shows the case
where it follows the Heston model.
For ResNet, we set $\sharp\Delta=1024$ and
$\Delta_1=\dots=\Delta_{1024}=1/1024$;
for NVnet $\sharp\Delta=4$ and $\Delta_1=\dots=\Delta_4=1/4$.
\subsubsection{Black--Scholes--Merton model case}
\begin{figure}
  \captionsetup{justification=centering}
  \includegraphics[width=14cm]{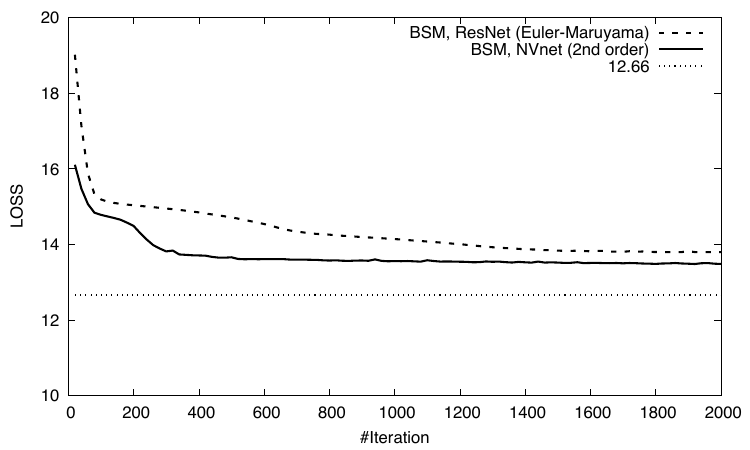}
  \captionof{figure}{ResNet vs NVnet, Black--Scholes--Merton model}
  \label{fig:BS}
\end{figure}

As demonstrated in Figure~\ref{fig:BS},
the ResNet combined with the Euler---Maruyama case,
\mbox{i.e.} the  $1$st order case, requires over $1500$ learning iterations
before the improvement due to learning becomes indistinguishable,
whereas the NVnet combined with the high-order discretisation case,
\mbox{i.e.} the $2$nd order case,
achieves the same outcome
after $250$ learning iterations.
It can also be seen that the optimisation results obtained by learning
are significantly better with NVnet than with ResNet, \mbox{i.e.} a lower
LOSS is achieved.
\par
If the price of the underlying asset follows the Black-Scholes-Merton
model, there is another well-known method of calculating the price of
American options using recombining binary trees~\cite{hull:2000book}.
The price calculated by this method is $12.66$
Figure~\ref{fig:BS} also shows prices calculated using this method
for reference.
It should be recalled that, as noted at the end of
subsection~\ref{ss:CommonParameters},
in these calculations, the LOSS coincides with the price itself
that is being sought.
\subsubsection{Heston model case}\label{ss:HestonCase}
\begin{figure}
  \captionsetup{justification=centering}
  \includegraphics[width=14cm]{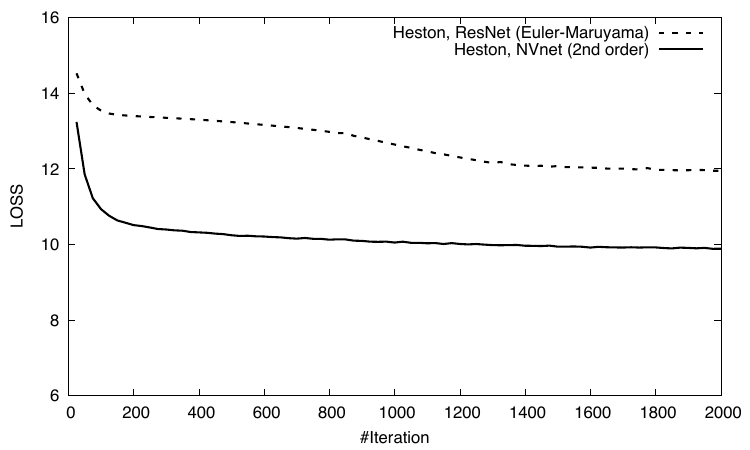}
  \captionof{figure}{ResNet vs NVnet, Heston model}\label{fig:HESTON}
\end{figure}

For the case of the Heston model, Figure~\ref{fig:HESTON} shows that
even with $2000$ learning iterations, both ResNet and NVnet still
improve slightly through learning. Otherwise, the trend is the same as
for the Black-Scholes-Merton model described above, but the difference
between ResNet and NVnet is even more pronounced.
In particular, the difference in learning speed---i.e. the rate of LOSS
reduction---is even greater: NVnet achieves an optimisation result in
just $50$ learning iterations---something that ResNet is unable to reach
even after extensive training.
\subsection{Discussions}\label{ss:discussion}
This series of numerical experiments shows that the concept of
high-order deep neural SDE networks, proposed from a purely
mathematical point of view, and its enabling architecture, NVnet, are
indeed effective in learning martingales. It performs significantly
better than the first-order network architecture, ResNet, both
in terms of the level of optimisation and the speed of learning.
\par
The objective of this numerical experiment is to ascertain whether the
optimisation process occurs as predicted. To this end, the
implementation utilises libraries such as TensorFlow in their current
state and without additional modifications, and crucial evaluations
from an engineering perspective, including the assessment of the
accuracy of the optimisation and the speed of learning, have not yet
been conducted.
\subsubsection{Immediate tasks}\label{ss:imediate}
In particular, the following two issues are immediately obvious and need to be investigated further: 
(1) that in the case of the Black--Scholes--Merton model, the values obtained from the optimisation do not seem to reach those obtained from the binomial tree.
(2) in the case of the Heston model, the number of training iterations does not seem to be sufficient, as described in subsection~\ref{ss:HestonCase}.
For the first one, it is assumed that the optimisation was not sufficient because it was implemented to work for the time being, without sufficiently investigating the implementation method of the vector field.
For the second, it is due to excessive memory consumption, also caused by the makeshift implementation, and it is essential to examine the design from an engineering perspective next time.
\subsubsection{NNnet}\label{ss:NNnet}
It is worth implementing NNnet and performing numerical verification in the same way as we do with NVnet. This is because, although both NVnet and NNnet are second-order SDE networks, their structures are completely different, and if they perform as well as NVnet, it will confirm the usefulness of the approach of learning martingales with higher-order NNs.
We are currently in the process of implementing NNnet.
\subsubsection{Monte Carlo and quasi-Monte Carlo free algorithms}
\label{ss:qMC-MC-free}
As stated in subsection~\ref{ss:order-of-discretisation}, the study
does not take into account any integration error.
This is because option pricing under the two asset models considered
here has already been examined in earlier
works~\cite{NinomiyaVictoir:2005, NinomiyaNinomiya:2007,doi:10.1080/1350486X.2019.1637268},
and integration errors are found to be negligible when using comparable sample
sizes with quasi-Monte Carlo.
\par
In general, however, it is worth considering algorithms that are free from
both Monte Carlo and quasi-Monte Carlo methods.
These approaches pose challenges when reproducibility, explainability,
and result verification are of concern.
Although quasi-Monte Carlo is reproducible in a strict sense,
it does not offer guarantees in practice, much like Monte Carlo.
\par
In this context, the high-order recombination measure method,
which the authors and others have been developing in recent years,
is considered promising.
Initially proposed in~\cite{litterer2012}, the method has since been
investigated in~\cite{NinomiyaShinozaki:york2023},
\cite{NinomiyaShinozaki:2021},
and \cite{ninomiya2025highorderrecombinationalgorithmweak},
where algorithmic implementations and applications to the weak approximation
of SDEs in finance have demonstrated its practical value.
\input{ninomiya_ma_paper_2024.bbl}

\end{document}

%% file: RK_5.latex
\setlength{\unitlength}{1579sp}%
\begin{picture}(11055,8316)(886,-8194)
\put(8949,-7649){\makebox(0,0)[lb]{\smash{\fontsize{8}{9.6}\usefont{T1}{ptm}{m}{it}{\color[rgb]{0,0,0}{}}
}}}
\put(5270,-2142){\makebox(0,0)[lb]{\smash{\fontsize{5}{6}\usefont{T1}{ptm}{m}{n}{\color[rgb]{0,0,0}+}%
}}}
{\color[rgb]{0,0,0}\thicklines
\put(5026,-3436){\circle{318}}
}%
\put(4970,-3492){\makebox(0,0)[lb]{\smash{\fontsize{5}{6}\usefont{T1}{ptm}{m}{n}{\color[rgb]{0,0,0}+}%
}}}
{\color[rgb]{0,0,0}\put(4576,-5086){\circle{318}}
}%
\put(4520,-5142){\makebox(0,0)[lb]{\smash{\fontsize{5}{6}\usefont{T1}{ptm}{m}{n}{\color[rgb]{0,0,0}+}%
}}}
{\color[rgb]{0,0,0}\put(3676,-6436){\circle{318}}
}%
\put(3620,-6492){\makebox(0,0)[lb]{\smash{\fontsize{5}{6}\usefont{T1}{ptm}{m}{n}{\color[rgb]{0,0,0}+}%
}}}
{\color[rgb]{0,0,0}\put(4576,-7786){\circle{318}}
}%
\put(4520,-7842){\makebox(0,0)[lb]{\smash{\fontsize{5}{6}\usefont{T1}{ptm}{m}{n}{\color[rgb]{0,0,0}+}%
}}}
{\color[rgb]{0,0,0}\put(9901,-4186){\circle{318}}
}%
\put(9845,-4242){\makebox(0,0)[lb]{\smash{\fontsize{5}{6}\usefont{T1}{ptm}{m}{n}{\color[rgb]{0,0,0}+}%
}}}
\put(6301,-811){\makebox(0,0)[lb]{\smash{\fontsize{10}{12}\usefont{T1}{ptm}{b}{it}{\color[rgb]{0,0,0}$Z_1$}%
}}}
\put(6526,-886){\makebox(0,0)[lb]{\smash{\fontsize{7}{8.4}\usefont{T1}{ptm}{b}{it}{\color[rgb]{0,0,0}{}}
}}}
\put(6226,-6511){\makebox(0,0)[lb]{\smash{\fontsize{10}{12}\usefont{T1}{ptm}{b}{it}{\color[rgb]{0,0,0}$Z_5$}%
}}}
\put(6451,-6586){\makebox(0,0)[lb]{\smash{\fontsize{8}{9.6}\usefont{T1}{ptm}{b}{it}{\color[rgb]{0,0,0}{}}
}}}
\put(6301,-3511){\makebox(0,0)[lb]{\smash{\fontsize{10}{12}\usefont{T1}{ptm}{b}{it}{\color[rgb]{0,0,0}$Z_3$}%
}}}
\put(6526,-3586){\makebox(0,0)[lb]{\smash{\fontsize{8}{9.6}\usefont{T1}{ptm}{b}{it}{\color[rgb]{0,0,0}{}} 
}}}
\put(6301,-5236){\makebox(0,0)[lb]{\smash{\fontsize{10}{12}\usefont{T1}{ptm}{b}{it}{\color[rgb]{0,0,0}$Z_4$}%
}}}
\put(6526,-5311){\makebox(0,0)[lb]{\smash{\fontsize{8}{9.6}\usefont{T1}{ptm}{b}{it}{\color[rgb]{0,0,0}{}}
}}}
\put(6226,-7861){\makebox(0,0)[lb]{\smash{\fontsize{10}{12}\usefont{T1}{ptm}{b}{it}{\color[rgb]{0,0,0}$Z_6$}%
}}}
\put(6451,-7936){\makebox(0,0)[lb]{\smash{\fontsize{8}{9.6}\usefont{T1}{ptm}{b}{it}{\color[rgb]{0,0,0}{}}
}}}
\put(6526,-2311){\makebox(0,0)[lb]{\smash{\fontsize{8}{9.6}\usefont{T1}{ptm}{b}{it}{\color[rgb]{0,0,0}{}}
}}}
\put(6301,-2236){\makebox(0,0)[lb]{\smash{\fontsize{10}{12}\usefont{T1}{ptm}{b}{it}{\color[rgb]{0,0,0}$Z_2$}%
}}}
\put(8251,-3211){\makebox(0,0)[lb]{\smash{\fontsize{10}{12}\usefont{T1}{ptm}{m}{it}{\color[rgb]{0,0,0}$\Delta_k b_3$}%
}}}
\put(8551,-3361){\makebox(0,0)[lb]{\smash{\fontsize{8}{9.6}\usefont{T1}{ptm}{m}{it}{\color[rgb]{0,0,0}{}}%
}}}
\put(8251,-6136){\makebox(0,0)[lb]{\smash{\fontsize{10}{12}\usefont{T1}{ptm}{m}{it}{\color[rgb]{0,0,0}$\Delta_k b_5$}%
}}}
\put(8551,-6286){\makebox(0,0)[lb]{\smash{\fontsize{8}{9.6}\usefont{T1}{ptm}{m}{it}{\color[rgb]{0,0,0}{}}
}}}
\put(8326,-7486){\makebox(0,0)[lb]{\smash{\fontsize{10}{12}\usefont{T1}{ptm}{m}{it}{\color[rgb]{0,0,0}$\Delta_k b_6$}%
}}}
\put(8626,-7636){\makebox(0,0)[lb]{\smash{\fontsize{8}{9.6}\usefont{T1}{ptm}{m}{it}{\color[rgb]{0,0,0}{}}%
}}}
\put(8176,-4711){\makebox(0,0)[lb]{\smash{\fontsize{10}{12}\usefont{T1}{ptm}{m}{it}{\color[rgb]{0,0,0}$\Delta_k b_4$}%
}}}
\put(8476,-4861){\makebox(0,0)[lb]{\smash{\fontsize{8}{9.6}\usefont{T1}{ptm}{m}{it}{\color[rgb]{0,0,0}{}}%
}}}
\put(8251,-1186){\makebox(0,0)[lb]{\smash{\fontsize{10}{12}\usefont{T1}{ptm}{m}{it}{\color[rgb]{0,0,0}$\Delta_k b_1$}%
}}}
\put(8551,-1336){\makebox(0,0)[lb]{\smash{\fontsize{8}{9.6}\usefont{T1}{ptm}{m}{it}{\color[rgb]{0,0,0}{}}
}}}
\put(901,-4261){\makebox(0,0)[lb]{\smash{\fontsize{10}{12}\usefont{T1}{ptm}{m}{it}{\color[rgb]{0,0,0}$x_k$}%
}}}
\put(1051,-4336){\makebox(0,0)[lb]{\smash{\fontsize{8}{9.6}\usefont{T1}{ptm}{m}{it}{\color[rgb]{0,0,0}{}}%
}}}
\put(11701,-4261){\makebox(0,0)[lb]{\smash{\fontsize{10}{12}\usefont{T1}{ptm}{m}{it}{\color[rgb]{0,0,0}$x_{k+1}$}%
}}}
\put(11926,-4336){\makebox(0,0)[lb]{\smash{\fontsize{8}{9.6}\usefont{T1}{ptm}{m}{it}{\color[rgb]{0,0,0}{}}%
}}}
{\color[rgb]{1,1,1}\put(5022,-2761){\circle*{92}}
}%
{\color[rgb]{1,1,1}\put(5018,-2761){\circle*{92}}
}%
{\color[rgb]{1,1,1}\put(5022,-2765){\circle*{106}}
}%
{\color[rgb]{0,0,0}\put(5476,-2086){\vector( 1, 0){675}}
}%
{\color[rgb]{0,0,0}\put(5176,-3436){\vector( 1, 0){975}}
}%
{\color[rgb]{0,0,0}\put(4726,-5086){\vector( 1, 0){1425}}
}%
{\color[rgb]{0,0,0}\put(3826,-6436){\vector( 1, 0){2325}}
}%
{\color[rgb]{0,0,0}\put(4726,-7786){\vector( 1, 0){1425}}
}%
{\color[rgb]{0,0,0}\put(6151,-3811){\framebox(600,750){}}
}%
{\color[rgb]{0,0,0}\put(6151,-5461){\framebox(600,750){}}
}%
{\color[rgb]{0,0,0}\put(6151,-8161){\framebox(600,750){}}
}%
{\color[rgb]{0,0,0}\put(1651,-961){\line( 0, 1){750}}
\multiput(1651,-211)(15.00000,15.00000){21}{\makebox(16.6667,25.0000){\small.}}
\put(1951, 89){\line( 1, 0){6750}}
\put(8701, 89){\line( 6,-5){1224.590}}
\put(9901,-961){\vector( 0,-1){3000}}
}%
{\color[rgb]{0,0,0}\put(10051,-4186){\vector( 1, 0){1650}}
}%
{\color[rgb]{0,0,0}\put(6751,-661){\line( 1, 0){1950}}
\put(8701,-661){\line( 5,-4){750}}
\put(9451,-1261){\line( 0,-1){2400}}
\multiput(9451,-3661)(15.00000,-15.00000){21}{\makebox(16.6667,25.0000){\small.}}
\put(9751,-3961){\vector( 1,-1){0}}
}%
{\color[rgb]{0,0,0}\put(1351,-4261){\line( 1, 0){300}}
\put(1651,-4261){\line( 0, 1){3300}}
\multiput(1651,-961)(15.00000,15.00000){21}{\makebox(16.6667,25.0000){\small.}}
\put(1951,-661){\vector( 1, 0){4125}}
}%
{\color[rgb]{0,0,0}\put(7651,-661){\line( 0,-1){600}}
\multiput(7651,-1261)(-15.00000,-15.00000){11}{\makebox(16.6667,25.0000){\small.}}
\put(7501,-1411){\line(-1, 0){5100}}
\multiput(2401,-1411)(-15.00000,-15.00000){11}{\makebox(16.6667,25.0000){\small.}}
\put(2251,-1561){\line( 0,-1){4350}}
\multiput(2251,-5911)(15.00000,-15.00000){11}{\makebox(16.6667,25.0000){\small.}}
\put(2401,-6061){\line( 1, 0){750}}
\multiput(3151,-6061)(18.75000,-9.37500){25}{\makebox(16.6667,25.0000){\small.}}
\put(3601,-6286){\vector( 2,-1){0}}
}%
{\color[rgb]{0,0,0}\multiput(5476,-1411)(-15.00000,-15.00000){11}{\makebox(16.6667,25.0000){\small.}}
\put(5326,-1561){\vector( 0,-1){375}}
}%
{\color[rgb]{0,0,0}\put(6751,-2086){\line( 1, 0){750}}
\multiput(7501,-2086)(15.00000,-15.00000){11}{\makebox(16.6667,25.0000){\small.}}
\put(7651,-2236){\line( 0,-1){375}}
\multiput(7651,-2611)(-15.00000,-15.00000){11}{\makebox(16.6667,25.0000){\small.}}
\put(7501,-2761){\line(-1, 0){2400}}
}%
{\color[rgb]{0,0,0}\put(4951,-2761){\line(-1, 0){2625}}
}%
{\color[rgb]{0,0,0}\multiput(5701,-2761)(-15.00000,-15.00000){36}{\makebox(16.6667,25.0000){\small.}}
\put(5176,-3286){\vector(-1,-1){0}}
}%
{\color[rgb]{0,0,0}\multiput(3601,-6211)(-15.00000,15.00000){31}{\makebox(16.6667,25.0000){\small.}}
\put(3601,-6211){\vector( 1,-1){0}}
\put(3151,-5761){\line(-1, 0){600}}
\multiput(2551,-5761)(-15.00000,15.00000){11}{\makebox(16.6667,25.0000){\small.}}
\put(2401,-5611){\line( 0, 1){2700}}
\multiput(2401,-2911)(15.00000,15.00000){11}{\makebox(16.6667,25.0000){\small.}}
}%
{\color[rgb]{0,0,0}\put(2476,-3436){\vector( 1, 0){2400}}
}%
{\color[rgb]{0,0,0}\put(2326,-3436){\makebox(16.6667,25.0000){\normalsize.}}
}%
{\color[rgb]{0,0,0}\put(2176,-3436){\line(-1, 0){375}}
\multiput(1801,-3436)(-15.00000,-15.00000){11}{\makebox(16.6667,25.0000){\small.}}
}%
{\color[rgb]{0,0,0}\put(8860,-3424){\vector( 4,-3){816}}
\put(8851,-3436){\line(-1, 0){2100}}
}%
{\color[rgb]{0,0,0}\put(4576,-4261){\vector( 0,-1){675}}
\multiput(4576,-4261)(15.00000,15.00000){11}{\makebox(16.6667,25.0000){\small.}}
\put(4726,-4111){\line( 1, 0){2775}}
\multiput(7501,-4111)(15.00000,15.00000){11}{\makebox(16.6667,25.0000){\small.}}
\put(7651,-3961){\line( 0, 1){375}}
\multiput(7651,-3586)(-15.00000,15.00000){11}{\makebox(16.6667,25.0000){\small.}}
}%
{\color[rgb]{0,0,0}\multiput(4501,-7636)(-15.00000,15.00000){16}{\makebox(16.6667,25.0000){\small.}}
\put(4501,-7636){\vector( 1,-1){0}}
\put(4276,-7411){\line( 0, 1){900}}
}%
{\color[rgb]{0,0,0}\put(4276,-6361){\line( 0, 1){150}}
\put(4276,-6211){\line( 1, 1){600}}
\put(4876,-5611){\line( 0, 1){450}}
}%
{\color[rgb]{0,0,0}\put(4876,-5011){\line( 0, 1){750}}
\multiput(4876,-4261)(15.00000,15.00000){11}{\makebox(16.6667,25.0000){\small.}}
}%
{\color[rgb]{0,0,0}\put(3676,-4261){\vector( 0,-1){1950}}
\multiput(3676,-4261)(15.00000,15.00000){11}{\makebox(16.6667,25.0000){\small.}}
\put(3826,-4111){\line( 1, 0){900}}
}%
{\color[rgb]{0,0,0}\put(2101,-7786){\vector( 1, 0){2250}}
\multiput(2101,-7786)(-15.00000,15.00000){31}{\makebox(16.6667,25.0000){\small.}}
\put(1651,-7336){\line( 0, 1){3075}}
}%
{\color[rgb]{0,0,0}\put(3751,-5086){\vector( 1, 0){675}}
}%
{\color[rgb]{0,0,0}\put(3601,-5086){\line(-1, 0){1125}}
}%
{\color[rgb]{0,0,0}\put(2326,-5086){\makebox(16.6667,25.0000){\normalsize.}}
}%
{\color[rgb]{0,0,0}\put(2176,-5086){\line(-1, 0){375}}
\multiput(1801,-5086)(-15.00000,15.00000){11}{\makebox(16.6667,25.0000){\small.}}
}%
{\color[rgb]{0,0,0}\put(3751,-7411){\vector( 2,-1){600}}
\put(3751,-7411){\line(-1, 0){1500}}
\multiput(2251,-7411)(-15.00000,15.00000){21}{\makebox(16.6667,25.0000){\small.}}
\put(1951,-7111){\line( 0, 1){1950}}
}%
{\color[rgb]{0,0,0}\put(1951,-5011){\line( 0, 1){1500}}
}%
{\color[rgb]{0,0,0}\put(2176,-2761){\line(-1, 0){150}}
\multiput(2026,-2761)(-15.00000,-15.00000){6}{\makebox(16.6667,25.0000){\small.}}
\put(1951,-2836){\line( 0,-1){525}}
}%
{\color[rgb]{0,0,0}\put(2026,-6436){\vector( 1, 0){1500}}
}%
{\color[rgb]{0,0,0}\put(1876,-6436){\line(-1, 0){ 75}}
\multiput(1801,-6436)(-15.00000,15.00000){11}{\makebox(16.6667,25.0000){\small.}}
}%
{\color[rgb]{0,0,0}\put(4576,-6811){\vector( 0,-1){825}}
\multiput(4576,-6811)(15.00000,15.00000){11}{\makebox(16.6667,25.0000){\small.}}
\put(4726,-6661){\line( 0, 1){150}}
}%
{\color[rgb]{0,0,0}\put(4726,-6361){\line( 0, 1){150}}
\multiput(4726,-6211)(15.00000,15.00000){31}{\makebox(16.6667,25.0000){\small.}}
\put(5176,-5761){\line( 1, 0){2325}}
\multiput(7501,-5761)(15.00000,15.00000){11}{\makebox(16.6667,25.0000){\small.}}
\put(7651,-5611){\line( 0, 1){375}}
\multiput(7651,-5236)(-15.00000,15.00000){11}{\makebox(16.6667,25.0000){\small.}}
\put(7501,-5086){\line(-1, 0){750}}
}%
{\color[rgb]{0,0,0}\put(5176,-5761){\line(-1, 0){375}}
}%
{\color[rgb]{0,0,0}\put(4651,-5761){\line(-1, 0){450}}
\multiput(4201,-5761)(-12.54545,-16.72727){34}{\makebox(16.6667,25.0000){\small.}}
\put(3787,-6313){\vector(-3,-4){0}}
}%
{\color[rgb]{0,0,0}\put(2326,-2086){\vector( 1, 0){2850}}
}%
{\color[rgb]{0,0,0}\put(2176,-2086){\line(-1, 0){375}}
\multiput(1801,-2086)(-15.00000,-15.00000){11}{\makebox(16.6667,25.0000){\small.}}
}%
{\color[rgb]{0,0,0}\put(4951,-2761){\line( 1, 0){150}}
}%
{\color[rgb]{0,0,0}\multiput(4876,-3286)(-18.17308,12.11538){21}{\makebox(16.6667,25.0000){\small.}}
\put(4876,-3286){\vector( 3,-2){0}}
\put(4501,-3061){\line( 0, 1){225}}
}%
{\color[rgb]{0,0,0}\put(4501,-2686){\line( 0, 1){525}}
}%
{\color[rgb]{0,0,0}\put(4501,-2011){\line( 0, 1){450}}
\multiput(4501,-1561)(15.00000,15.00000){11}{\makebox(16.6667,25.0000){\small.}}
}%
{\color[rgb]{0,0,0}\put(9901,-7411){\vector( 0, 1){3000}}
\multiput(9901,-7411)(-16.07143,-13.39286){29}{\makebox(16.6667,25.0000){\small.}}
\put(9451,-7786){\line(-1, 0){2700}}
}%
{\color[rgb]{0,0,0}\multiput(9826,-4411)(-6.81818,-20.45455){23}{\makebox(16.6667,25.0000){\small.}}
\put(9826,-4411){\vector( 1, 3){0}}
\put(9676,-4861){\line( 0,-1){1350}}
\multiput(9676,-6211)(-16.66667,-12.50000){19}{\makebox(16.6667,25.0000){\small.}}
\put(9376,-6436){\line(-1, 0){1875}}
}%
{\color[rgb]{0,0,0}\put(8776,-5086){\vector( 4, 3){900}}
\put(8776,-5086){\line(-1, 0){1275}}
}%
\thinlines
{\color[rgb]{0,0,0}\put(6151,-1081){\framebox(600,750){}}
}%
{\color[rgb]{0,0,0}\put(6136,-3811){\framebox(600,750){}}
}%
{\color[rgb]{0,0,0}\put(6151,-5461){\framebox(600,750){}}
}%
{\color[rgb]{0,0,0}\put(6151,-8161){\framebox(600,750){}}
}%
\thicklines
{\color[rgb]{0,0,0}\put(6151,-2461){\framebox(600,750){}}
}%
{\color[rgb]{0,0,0}\put(6151,-6811){\framebox(600,750){}}
}%
\thinlines
{\color[rgb]{0,0,0}\put(6151,-6811){\framebox(600,750){}}
}%
\thicklines
{\color[rgb]{0,0,0}\multiput(4651,-7636)(12.54545,16.72727){34}{\makebox(16.6667,25.0000){\small.}}
\put(4651,-7636){\vector(-3,-4){0}}
\put(5101,-7111){\line( 1, 0){2400}}
\multiput(7501,-7111)(15.00000,15.00000){11}{\makebox(16.6667,25.0000){\small.}}
\put(7651,-6961){\line( 0, 1){375}}
\multiput(7651,-6586)(-15.00000,15.00000){11}{\makebox(16.6667,25.0000){\small.}}
\put(7501,-6436){\line(-1, 0){750}}
}%
{\color[rgb]{0,0,0}\put(6151,-1081){\framebox(593,743){}}
}%
\put(8964,-1321){\makebox(0,0)[lb]{\smash{\fontsize{8}{9.6}\usefont{T1}{ptm}{m}{it}{\color[rgb]{0,0,0}{}}
}}}
\put(8963,-3391){\makebox(0,0)[lb]{\smash{\fontsize{8}{9.6}\usefont{T1}{ptm}{m}{it}{\color[rgb]{0,0,0}{}}
}}}
\put(3939,-2543){\makebox(0,0)[lb]{\smash{\fontsize{10}{12}\usefont{T1}{ptm}{m}{it}{\color[rgb]{0,0,0}a}%
}}}
\put(5514,-1831){\makebox(0,0)[lb]{\smash{\fontsize{10}{12}\usefont{T1}{ptm}{m}{it}{\color[rgb]{0,0,0}a}%
}}}
\put(5551,-3195){\makebox(0,0)[lb]{\smash{\fontsize{10}{12}\usefont{T1}{ptm}{m}{it}{\color[rgb]{0,0,0}a}%
}}}
\put(4710,-6968){\makebox(0,0)[lb]{\smash{\fontsize{10}{12}\usefont{T1}{ptm}{m}{it}{\color[rgb]{0,0,0}a}%
}}}
\put(2596,-1718){\makebox(0,0)[lb]{\smash{\fontsize{10}{12}\usefont{T1}{ptm}{m}{it}{\color[rgb]{0,0,0}a}%
}}}
\put(5709,-1951){\makebox(0,0)[lb]{\smash{\fontsize{8}{9.6}\usefont{T1}{ptm}{m}{it}{\color[rgb]{0,0,0}21}%
}}}
\put(3968,-4635){\makebox(0,0)[lb]{\smash{\fontsize{10}{12}\usefont{T1}{ptm}{m}{it}{\color[rgb]{0,0,0}a}%
}}}
\put(4164,-4756){\makebox(0,0)[lb]{\smash{\fontsize{8}{9.6}\usefont{T1}{ptm}{m}{it}{\color[rgb]{0,0,0}43}%
}}}
\put(3053,-4643){\makebox(0,0)[lb]{\smash{\fontsize{10}{12}\usefont{T1}{ptm}{m}{it}{\color[rgb]{0,0,0}a}%
}}}
\put(3241,-4757){\makebox(0,0)[lb]{\smash{\fontsize{8}{9.6}\usefont{T1}{ptm}{m}{it}{\color[rgb]{0,0,0}53}%
}}}
\put(4119,-2656){\makebox(0,0)[lb]{\smash{\fontsize{8}{9.6}\usefont{T1}{ptm}{m}{it}{\color[rgb]{0,0,0}31}%
}}}
\put(5739,-3301){\makebox(0,0)[lb]{\smash{\fontsize{8}{9.6}\usefont{T1}{ptm}{m}{it}{\color[rgb]{0,0,0}32}%
}}}
\put(2619,-5558){\makebox(0,0)[lb]{\smash{\fontsize{10}{12}\usefont{T1}{ptm}{m}{it}{\color[rgb]{0,0,0}a}%
}}}
\put(3998,-5618){\makebox(0,0)[lb]{\smash{\fontsize{10}{12}\usefont{T1}{ptm}{m}{it}{\color[rgb]{0,0,0}a}%
}}}
\put(4209,-5701){\makebox(0,0)[lb]{\smash{\fontsize{8}{9.6}\usefont{T1}{ptm}{m}{it}{\color[rgb]{0,0,0}54}%
}}}
\put(2799,-5679){\makebox(0,0)[lb]{\smash{\fontsize{8}{9.6}\usefont{T1}{ptm}{m}{it}{\color[rgb]{0,0,0}52}%
}}}
\put(2791,-1816){\makebox(0,0)[lb]{\smash{\fontsize{8}{9.6}\usefont{T1}{ptm}{m}{it}{\color[rgb]{0,0,0}51}%
}}}
\put(4921,-7089){\makebox(0,0)[lb]{\smash{\fontsize{8}{9.6}\usefont{T1}{ptm}{m}{it}{\color[rgb]{0,0,0}64}%
}}}
\put(5071,-7537){\makebox(0,0)[lb]{\smash{\fontsize{10}{12}\usefont{T1}{ptm}{m}{it}{\color[rgb]{0,0,0}a}%
}}}
\put(5281,-7659){\makebox(0,0)[lb]{\smash{\fontsize{8}{9.6}\usefont{T1}{ptm}{m}{it}{\color[rgb]{0,0,0}65}%
}}}
\put(3676,-7163){\makebox(0,0)[lb]{\smash{\fontsize{10}{12}\usefont{T1}{ptm}{m}{it}{\color[rgb]{0,0,0}a}%
}}}
\put(3864,-7314){\makebox(0,0)[lb]{\smash{\fontsize{8}{9.6}\usefont{T1}{ptm}{m}{it}{\color[rgb]{0,0,0}63}%
}}}
\put(2490,-7186){\makebox(0,0)[lb]{\smash{\fontsize{10}{12}\usefont{T1}{ptm}{m}{it}{\color[rgb]{0,0,0}a}%
}}}
\put(2709,-7299){\makebox(0,0)[lb]{\smash{\fontsize{8}{9.6}\usefont{T1}{ptm}{m}{it}{\color[rgb]{0,0,0}62}%
}}}
\put(8882,-4854){\makebox(0,0)[lb]{\smash{\fontsize{8}{9.6}\usefont{T1}{ptm}{m}{it}{\color[rgb]{0,0,0}{}}
}}}
\put(8949,-6268){\makebox(0,0)[lb]{\smash{\fontsize{8}{9.6}\usefont{T1}{ptm}{m}{it}{\color[rgb]{0,0,0}{}}
}}}
{\color[rgb]{0,0,0}\put(5326,-2086){\circle{318}}
}%
\end{picture}%
%

%% file: ninomiya_ma_paper_2024.bbl
\providecommand{\bysame}{\leavevmode\hbox to3em{\hrulefill}\thinspace}
\providecommand{\MR}{\relax\ifhmode\unskip\space\fi MR }
\providecommand{\MRhref}[2]{%
  \href{http://www.ams.org/mathscinet-getitem?mr=#1}{#2}
}
\providecommand{\href}[2]{#2}